\documentclass{Interspeech}

\interspeechcameraready

\title{Speech Unlearning}

\author[]{Jiali}{Cheng}
\author[]{Hadi}{Amiri}

\affiliation{}{University of Massachusetts Lowell}{USA}
\email{\texttt{\{jiali\_cheng, hadi\_amiri\}@uml.edu}}

\usepackage{comment}
\usepackage{multirow}

\begin{document}

\maketitle

\begin{abstract}
    We introduce {\em machine unlearning for speech tasks}, a novel and underexplored research problem that aims to efficiently and effectively remove the influence of specific data from trained speech models without full retraining. This has important applications in privacy preservation, removal of outdated or noisy data, and bias mitigation. 
    While machine unlearning has been studied in computer vision and natural language processing, its application to speech is largely unexplored due to the high-dimensional, sequential, and speaker-dependent nature of speech data. 
    We define two fundamental speech unlearning tasks:
    \textit{sample unlearning}, which removes individual data points (e.g., a voice recording), and 
    \textit{class unlearning}, which removes an entire category (e.g., all data from a speaker), while preserving performance on the remaining data. %
    Experiments on keyword spotting and speaker identification demonstrate that unlearning speech data is significantly more challenging than unlearning image or text data. 
    We conclude with key future directions in this area, including structured training, robust evaluation, feature-level unlearning, broader applications, scalable methods, and adversarial robustness.
\end{abstract}
\section{Introduction}


Machine unlearning~\cite{bourtoule2021machine,Golatkar2020EternalSO} aims to remove specific data (e.g., training examples or individual classes) from a trained model while preserving performance on the remaining data. While progress has been made in computer vision~\cite{Golatkar2020EternalSO}, natural language processing~\cite{jia-etal-2024-soul}, and recommendation systems~\cite{10.1145/3485447.3511997}, its application to speech models is largely unexplored.

We introduce {\em speech unlearning}, which aims to remove a subset of training data and its influence from trained speech models. Unlike unlearning in other modalities, speech unlearning has unique challenges due to 
temporal dependencies and sequence complexity in speech models; 
entanglement of phonetic and speaker-related features; and 
the need for high precision in speech-based applications to preserve privacy. 

We define two core unlearning tasks in speech, both aim to remove specific information without requiring retraining:
{\em sample unlearning}, which removes specific data points (e.g., a voice recording) while preserving knowledge on other  data points, and 
{\em class unlearning}, which erases an entire category (e.g., a speaker or keyword) without affecting recognition of other categories, see~\S\ref{sec:core}.
Through case studies on keyword spotting and speaker identification tasks, we demonstrate 
the difficulty of disentangling learned speech representations and the trade-offs between unlearning effectiveness and model utility. 
We propose future research directions to overcome these challenges.
%
The key contributions of this work are:
\begin{itemize}
    \item we formally define the task of speech unlearning and highlight its importance and challenges,
    \item we conduct experiments on keyword spotting and speaker identification to quantify the difficulty of disentangling learned speech representations and the trade-offs between unlearning effectiveness and model utility, and 
    \item we outline open challenges and promising research directions to advance speech unlearning.
\end{itemize}


Experiments show that existing unlearning methods struggle to remove speech representations without harming overall model performance. In terms of sample unlearning, gradient ascent-based approaches degrade accuracy on both retained data and the data marked for deletion, while random labeling-based approaches fail to fully erase learned information. Class unlearning is even harder, as removing categories disrupts overall recognition. A structured forgetting approach improves unlearning efficiency, preserving retained data while better eliminating the forget set. These results highlight the need for more targeted, feature-aware unlearning strategies for speech models.

\begin{figure}
    \centering
    \includegraphics[width=0.99\linewidth]{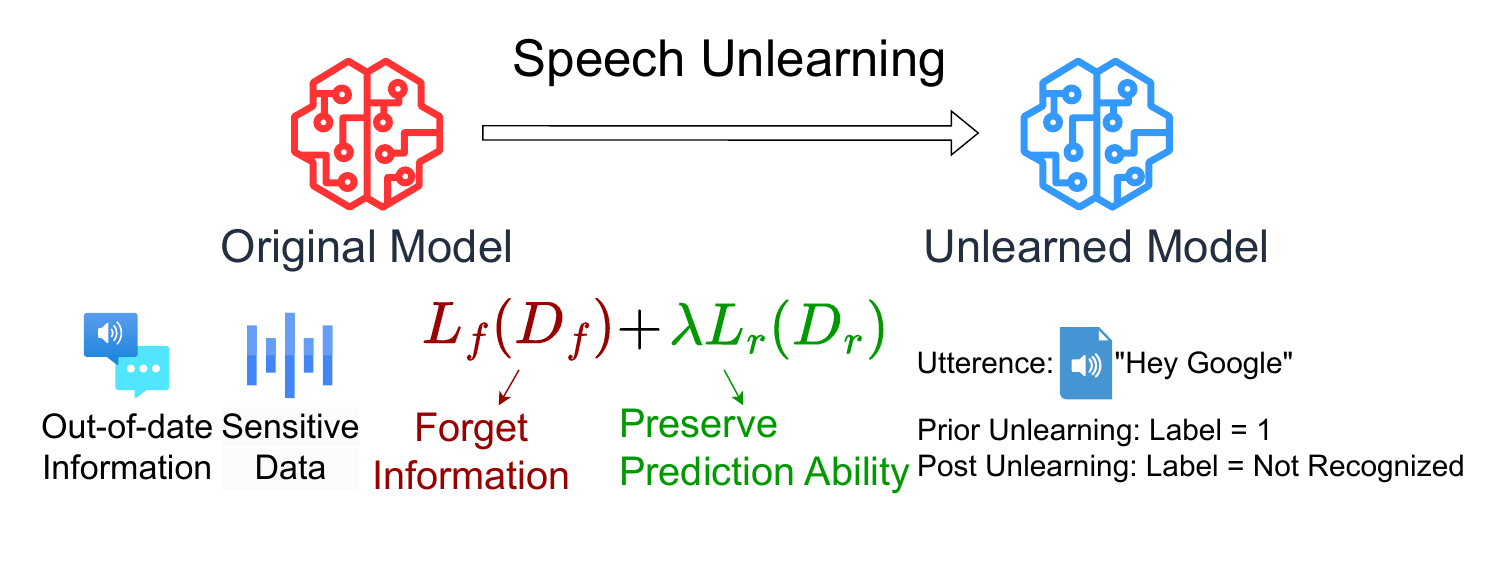}
    \vspace{-10pt}
    \caption{Illustration of Speech Unlearning.}    
    \label{fig:unlearn}
    \vspace{-15pt}
\end{figure}

\section{Related Work}

Machine unlearning was initially studied on vision tasks, specifically, image classification~\cite{bourtoule2021machine}. Existing methods approach unlearning with efficient retraining by dividing data into chunks and training a separate model on each chunk~\cite{Dukler_2023_ICCV,Lin_2023_CVPR}. Influence-based unlearning methods derives a one-shot update with theoretical guarantee, but usually limited to convex loss functions~\cite{suriyakumar2022algorithms,liu2023certified}
In addition, other vision tasks have been studied, including image retrieval~\cite{10.1145/3503161.3548378}, image-text classification~\cite{poppi2024removing}, text-to-image generation~\cite{zhang2024unlearncanvas} and image-to-image generation \cite{li2024machine}.
Recently, more attention has been given to unlearning on language tasks and Large Language Models (LLMs)~\cite{maini2024tofu}, where gradient ascent is a common technique~\cite{eldan2023whos}. Other methods include performing direct preference optimization with no positive examples~\cite{zhang2024negative}, deviating from original logits~\cite{ji2024reversing}, and mitigating highly memorized samples~\cite{kassem-etal-2023-preserving}.
%
Other works investigate unlearning on graphs~\cite{Chien2023EfficientMU,cheng2023gnndelete}, 
multimodal data~\cite{cheng2023multimodal},
regression task~\cite{pmlr-v202-tarun23a}, 
loss landscape of unlearning~\cite{cheng2025understanding},
traditional machine learning models~\cite{pan2023machine}, and 
continual learning setting~\cite{pmlr-v199-liu22a}. 
%


\section{Challenges of Speech Unlearning}





Speech unlearning presents unique challenges beyond those in other modalities due to the {\em multi-dimensional}, {\em sequential}, and {\em speaker-dependent} nature of speech data. In what follows, we describe some of these challenges:

\subsection{Temporal Dependencies and Sequence Complexity}
Unlike images or text data, speech data is a continuous and time-dependent signal, making precise removal of learned patterns challenging. For example, removing or altering a portion of the data could disrupt the entire sequence. 
In addition, the pronunciation of phonemes and the meaning of words can change depending on intonation, stress, and the surrounding context, which make fine-grained unlearning more complex. 
Furthermore, speech data can contain long-range phonetic and prosodic structures, and speech models encode information over extended time frames, both complicate targeted data removal.

\subsection{Representation Complexity}
Speech is represented as raw waveforms or spectrograms--high dimensional signals where small modifications can have significant effects. 
Unlike localized spatial features and discrete tokens in images and texts, speech data is continuous and lacks clear separation between elements, which makes selective removal non-trivial. In addition, removing certain phonetic patterns might alter linguistic features or affect speaker identify.

\subsection{Speaker-Specific Information and Personalization}
Speech contains rich personal identifiers such as pitch, tone, and speech patterns. Simply removing or modifying certain keywords does not erase the underlying speaker identity, which makes privacy-oriented unlearning a complex task~\cite{eldan2023whos,cheng2025tool}. 

\subsection{Multi-modal and multi-domain dependencies}
Speech models often interact with textual and visual data. For example, Whisper~\cite{radford2023robust} jointly learns acoustic and linguistic embeddings.  
Consequently, unlearning speech-related information may also necessitate the removal of associated text and image data, which further complicates the speech unlearning process~\cite{cheng24c_interspeech}.


\subsection{Interleaved Phonetic and Linguistic Information}
The same word can be pronounced differently across speakers, accents, and speaking conditions. Simply removing specific speech instances doesn’t necessarily remove all learned variations, as models can generalize across pronunciations.
In addition, {\em coarticulation}--the interaction of neighboring phonemes--complicates selective phoneme removal as it can introduce distortion or affect intelligibility.

\section{Machine Unlearning in Speech}\label{sec:prem}

\subsection{Problem Definition}
Given a speech model $f$ trained on dataset $D_{train}$, machine unlearning aims to remove the influence of a subset $D_f \in D_{train}$ (the forget set) from $f$ without full retraining, while preserving knowledge on the retain data $D_r = D_{train} \setminus D_f$. We term $f$ as the original model and $f'$ as the model post-unlearning.

\subsection{Core Speech Unlearning Tasks}\label{sec:core}
We define two fundamental speech unlearning tasks:
{\bf Sample Unlearning}: removing specific data points, e.g. a specific voice recording while preserving knowledge of the remaining data; and 
{\bf Class Unlearning}: removing an entire class or category, e.g. a specific speaker or keyword, while maintaining recognition accuracy on the remaining classes, e.g., preserving the ability to identify other speakers or unaffected keywords.

Both tasks are essential for protecting user privacy, 
legal compliance (GDPR and CCPA), and ethical AI practices. 
%
Class unlearning is particularly useful when removing outdated or sensitive content from speech models. For example, a company may need to phase out a wake word like ``Hey Google'' when introducing a new activation phrase or filter out offensive, deceptive, or harmful speech for safer interactions. Class unlearning is also crucial for security and compliance in voice authentication systems, where certain user data may need to be fully removed due to policy changes, security threats, or account deactivation, while keeping recognition accurate for other.


\subsection{A Unified Speech Unlearning Approach}
Speech unlearning can be formulated as an optimization problem, where the goal is to fine-tune the original model $f$ to remove knowledge of the forget set $D_f$ while preserving performance on the retain set $D_r$. This can be achieved by optimizing a forgetting loss $L_f$ to encourage unlearning $D_f$, and a retaining loss $L_r$ to enforce retention of $D_r$~\cite{jia-etal-2024-soul,cheng2024mubench,zhang2024towards,fan2024salun}:
\begin{equation}
    f' = \arg\min_{\theta'} L_{f} (D_f) + \lambda L_{r} (D_r),
\end{equation} 
where $\lambda$ is a hyperparameter that controls the trade-off between the forgetting and retention objectives. 
Below, we list several commonly-used unlearning methods within this framework, along with their choices of $L_f$ and $L_r$:
\begin{itemize}
    \item \textbf{Gradient Ascent:} 
    GradAscent~\cite{Golatkar2020EternalSO} applies gradient ascent on $D_f$ to remove learned representations. Specifically, $L_f$ is the negative of the task loss used for training the original model $f$, i.e. $L_f = -L_{\text{task}}$, and $\lambda = 0$. \\

    \item \textbf{Random Labeling:} 
    RandomLabel~\cite{Golatkar2020EternalSO} fine-tunes the original model on $D_f$ with corrupted labels and $D_r$. This process encourages the unlearned model to memorize randomized labels for unlearning. Here $L_r$ and $L_f$ are $L_{\mathrm{task}}$ and $\lambda = 1$. \\
    
    \item \textbf{Saliency Unlearning:}
    SalUn~\cite{fan2024salun} first finds parameter that are most salient to unlearning $D_f$ and then performs RandomLabel~\cite{Golatkar2020EternalSO} by only updating the salient parameters. \\

    \item \textbf{SCRUB:} 
    SCRUB~\cite{zhang2024towards} first maximizes error on $D_f$ with GradAscent. After that, it minimizes the error as well as the embedding differences between $f'$ and $f$ on $D_r$. \\
    
    \item  \textbf{Bad Teaching:}
    Bad-T~\cite{Chundawat_Tarun_Mandal_Kankanhalli_2023} forces the unlearned model to mimic the original model on $D_r$ while mimicking an incompetent model (e.g. a randomly initialized model) on $D_f$. Here $\lambda = 1$, and $L_f$ and $L_r$ are defined as the KL-Divergence between prediction logits: $\mathbb{KL}(f'(D_f) || f_d(D_f))$ and $\mathbb{KL}(f'(D_r) || f(D_r))$ respectively, where $f_d$ is the incompetent model. \\
    
\end{itemize}

\subsection{Evaluation Metrics for Speech Unlearning}
An ideal unlearned model $f'$ should effectively remove the knowledge of $D_f$ from the original model $f$, while preserving its performance on the remaining data $D_r$. We propose to use two metrics for effective evaluation: 
{\bf Subset Performance}, which computes the performance of the unlearned model $f'$ on the forget set $D_f (\downarrow)$, retain set $D_r (\uparrow)$, and the original test set $D_t (\uparrow)$. Ideally, an unlearned model should perform at random chance on $D_f$ to confirm successful forgetting. Meanwhile, the performance on $D_r$ should be preserved as much as possible, and the overall model performance on the original test data $D_t$ should remain stable. 
{\bf Membership Inference Attack (MIA)}, which is a technique to determine whether a data point was involved in the training set of a given model. In the context of machine unlearning, MIA helps assess whether the unlearned model still retains traces of the forgot set $D_f$ after the unlearning process. If a MIA model classifies most samples of $D_f$ as non-members of the training set, it indicates that $f'$ has limited knowledge of $D_f$. The most commonly-used MIA approach is black-box MIA, where membership inference is performed based on the model's loss values on given samples.

\section{Experiments}

\subsection{Datasets} 
We consider two speech tasks using data from the Superb benchmark~\cite{yang21c_interspeech}: keyword spotting and speaker identification. 
{\bf Keyword Spotting:} 
the goal is to detect predefined keywords, such as ``Hey Siri'' or ``Okay Google,'' in speech recordings. 
In case of sample unlearning, user request the removal of their specific voice recordings from the training set. The model must efficiently and effectively remove the contribution of all voice recordings marked for deletion. 
In case of class unlearning, the model should forget how to recognize specific keywords (e.g., ``Okay Google'') while preserving the ability to recognize all remaining keywords. 
For keyword spotting, we choose Speech Commands~\cite{speechcommands}, which consists of 85K utterances of 12 keyword classes. 
{\bf Speaker Identification:} 
aims to recognize a speaker's identity based on their voice characteristics. 
Here, sample unlearning is concerned with removing the influence of a subset of speech samples from the trained model, while retaining the ability to identify other speeches accurately. 
For class unlearning, the model should forget a specific speaker while ensuring that other speakers are identifiable.
For speaker identification, we use the VoxCeleb1 dataset~\cite{NAGRANI2020101027}, which contains 148K utterances from 1,211 speakers. 

\begin{table}[t!]
  \caption{Speech unlearning on keyword spotting task.}
  \label{tab:ks}
  \centering
    \renewcommand{\arraystretch}{1}
    \resizebox{\linewidth}{!}{
      \begin{tabular}{ l|ccccc }
        \toprule
        \textbf{Methods} & $\mathbf{D_t} (\uparrow)$ & $\mathbf{D_f} (\downarrow)$ & $\mathbf{D_r} (\uparrow)$ & \textbf{MIA} $(\uparrow)$ & \textbf{Time (hrs)} $(\downarrow)$ \\
        \midrule
        Original   & 96.1 & 97.4 & 99.8 & 17.3 & NA \\
        \midrule
        \multicolumn{5}{l}{\emph{Sample Unlearning}} \\
        \midrule
        GradAscent &  7.1 & 12.3 & 52.5 & 47.3 & 0.3 \\
        RandLabel  & 90.4 & 95.1 & 99.8 & 39.2 & 1.2 \\
        Bad-T      &  8.6 & 10.7 & 72.7 & 52.4 & 1.1 \\
        SCRUB      &  8.5 & 11.2 & 72.2 & 52.5 & 0.9 \\
        SalUn      & 90.4 & 95.3 & 99.7 & 40.5 & 1.3 \\
        \midrule
        \multicolumn{5}{l}{\emph{Class Unlearning}} \\
        \midrule
        GradAscent &  3.8 &  6.2 & 23.5 & 42.1 & 0.8 \\
        RandLabel  & 57.3 & 24.2 & 63.1 & 44.7 & 2.6 \\
        Bad-T      & 41.7 & 53.2 & 46.6 & 51.3 & 2.7 \\
        SCRUB      & 33.7 & 25.8 & 29.4 & 47.1 & 2.5 \\
        SalUn      & 42.3 & 39.4 & 40.1 & 50.7 & 2.7 \\
        \bottomrule
  \end{tabular}
  }
\end{table}

\subsection{Experimental Setup}
We use accuracy as the evaluation metric. For each task, we first train a base model $f$ with hyperparameter optimization and select the best performing checkpoint as the starting point of unlearning. In addition, to ensure unlearning remains practical, we impose a time constraint so that the unlearning process does not exceed the time required for full retraining--otherwise, unlearning would lose its advantage.
We evaluate unlearning across several backbone architectures, including Whisper (Tiny, Base)~\cite{radford2023robust}, Wav2Vec 2.0 (Base, Large)~\cite{baevski2020wav2vec}, and HuBert (Base, Large, X-Large)~\cite{hsu2021hubert}. Since real-world speech applications may be deployed on various model architectures, we report average performance across these models for a more generalizable assessment. 
For each dataset, we randomly select 1-10\% of data as the forget set, increasing in 1\% increments. 
We repeat all experiments five times with different random seeds to account for any stochastic effects.


\subsection{Main Results}
Results in Tables~\ref{tab:ks}~and~\ref{tab:si} show significant challenges in achieving effective speech unlearning.
For sample unlearning, none of the methods successfully unlearn $D_f$ without substantially degrading performance on $D_t$ and $D_r$. GradAscent, Bad-T and SCRUB achieve good forgetting on $D_f$, but at the cost of severe performances deterioration across other subsets. This is because 
these methods lack mechanisms to maintain performance on $D_r$. 
Conversely, RandLabel and SalUn still retain substantial knowledge of $D_f$, suggeting that fine tuning on mislabeled $D_f$ is insufficient for true unlearning. This can potentially be due to high inter-frame correlations and temporal dependencies in speech data, which existing unlearning methods fail to address. 

For class unlearning, results show an even greater challenge. While GradAscent achieves the lowest $D_f$ scores, it significantly reducing accuracy across other subsets. Methods such as Bad-T and SCRUB achieve moderate forgetting but struggle to balance forgetting and knowledge retention. RandLabel and SalUn show higher retention of $D_r$ but fail to fully unlearn $D_f$, reinforcing the difficulty of selectively removing an entire class without disrupting the learned representation space.


Across both tasks, the MIA scores suggest that current unlearning techniques are not entirely effective at preventing data reconstruction, as unlearned samples remain detectable.

\begin{table}[t!]
  \caption{Speech unlearning on speaker identification task.}
  \label{tab:si}
  \centering
  \renewcommand{\arraystretch}{1}
  \resizebox{\linewidth}{!}{
      \begin{tabular}{ l|ccccc }
        \toprule
        \textbf{Methods} & $\mathbf{D_t} (\uparrow)$ & $\mathbf{D_f} (\downarrow)$ & $\mathbf{D_r} (\uparrow)$ & \textbf{MIA} $(\uparrow)$ & \textbf{Time (hrs)} $(\downarrow)$ \\
        \midrule
        Original   & 68.6 & 85.6 & 87.3 & 21.1 & NA \\
        \midrule
        \multicolumn{5}{l}{\emph{Sample Unlearning}} \\
        \midrule
        GradAscent &  9.2 & 11.3 & 15.1 & 42.6 & 2.4 \\
        RandLabel  & 59.5 & 33.4 & 42.7 & 55.3 & 6.7 \\
        Bad-T      & 52.7 & 49.5 & 55.6 & 51.4 & 6.5 \\
        SCRUB      & 47.3 & 49.5 & 49.8 & 50.7 & 6.3 \\
        SalUn      & 58.7 & 33.6 & 47.2 & 54.6 & 6.8 \\
        \midrule
        \multicolumn{5}{l}{\emph{Class Unlearning}} \\
        \midrule
        GradAscent & 10.7 & 14.5 & 17.1 & 50.2 &  4.7 \\
        RandLabel  & 52.3 & 40.9 & 52.7 & 53.0 & 10.1 \\
        Bad-T      & 42.5 & 41.4 & 48.0 & 51.3 &  9.8 \\
        SCRUB      & 37.4 & 39.3 & 40.2 & 48.5 & 10.5 \\
        SalUn      & 45.1 & 44.3 & 45.7 & 49.6 & 10.2 \\
        \bottomrule
      \end{tabular}
    }
\end{table}

\subsection{A Structured Forgetting Strategy}
The fundamental challenge in speech unlearning is targeted forgetting while preserving overall model utility. Inspired by curriculum learning (CL)~\cite{cl_bengio}, we analyzed the potential of dynamic sample re-weighting using SuperLoss~\cite{superloss}, which uses the following objective function with a closed form solution to infer sample weights: $\mathcal{L}_\lambda = (l_i - \tau)  \sigma_i + \lambda (\log{\sigma_i})^2,$
where, at each training iteration, $\tau$ is the moving average of loss to separate easy samples from hard ones, and the weight ($\sigma$) of each instance is computed as a transformation of its loss with respect to $\tau$. 
With SuperLoss, the overall unlearning performance on $D_f$ effectively decreased from 44.7 to 17.3--a significant improvement in unlearning. This suggest that structured forgetting strategies, such as curriculum-based sample prioritization, could be a promising direction for scalable and adaptive unlearning techniques in speech models.

\section{Future Directions}
Speech unlearning is still in its early stages with several open challenges and research opportunities. Building on our findings and analysis, we discuss key advancements needed to develop effective speech unlearning methods, establish stronger evaluation metrics, extend unlearning to broader speech applications, and address possible risks. By tackling these challenges, we can move toward scalable, privacy-preserving, and adaptable speech AI systems.

\subsection{New Evaluation Metrics and Guarantees}
Existing unlearning works mainly rely on subset performance metrics to assess unlearning approaches. These metrics may be insufficient because models may learn to suppress specific results instead of truly removing influences of data~\cite{hu2025jogging}. 
A more robust evaluation framework should incorporate: 
{\em privacy leakage analysis} to provide guarantees that unlearned samples cannot be reconstructed or inferred; 
{\em feature attribution audits} to verify that the key audio features associated with the forget set $D_f$ are no longer used; and 
{\em membership inference resistance} to ensure that the unlearned samples remain undetectable.
Establishing theoretical guarantees for unlearning in speech system is an open challenge. Future research should explore certifiable unlearning methods to provide verifiable removal of targeted speech data.


\subsection{Feature-Level Speech Unlearning}
Current unlearning methods mainly focus on removing entire samples or classes, but feature-level unlearning--removing specific data attributes while preserving general utility--is unexplored. We recommend the following tasks: 
{\em Voice Attribute Unlearning}: can we remove a speaker's voice characteristics (e.g. pitch, timbre, and cadence) while preserving linguistic content? This is crucial in applications where speaker anonymity is required, such as de-identification in medical transcription or anonymous voice assistants.
{\em Accent Unlearning}: can we unlearn specific accents while retaining speech recognition ability? Speech recognition systems may carry biases where some dialects or regional variations are overrepresented. Accent unlearning would allow models to forget a particular accent’s characteristics to prevent overfitting to dominant accents.
{\em Style Unlearning}: can a model forget prosodic or emotional speech patterns without affecting transcription accuracy?
In some applications, it is necessary to remove style variations while preserving meaning of the content, e.g., in removing emotional tone for legal transcriptions or neutralizing bias in sentiment analysis.

\subsection{Broader Speech Applications}
This work studies speech unlearning on keyword spotting and speaker identification. Speech unlearning can extend to more complex and multimodal tasks, such as: 
{\em Noise Source Unlearning}: forget specific types of background noise (e.g., office or street noise) or misannotated speech samples from ASR models to improve their generalization.
{\em Language Unlearning}: forget a specific language or dialect from a multilingual ASR model, e.g., to adapt models to specific regions or removing unsupported languages from commercial models.
{\em Synthetic Voice Unlearning}: forget synthetic or adversarially generated speech samples or deepfake voices.
{\em Unlearn Demographic Biases}: remove biases that make ASR more accurate for one demographic group over another.

\subsection{More Effective and Efficient Speech Unlearning}
Developing scalable and efficient unlearning techniques is crucial for real-world deployment. Lightweight approaches such as influence-based data removal, structured pruning, and memory-efficient fine-tuning may achieve targeted speech unlearning with minimal computational overhead. In addition, incremental unlearning strategies, where models progressively or sequentially forget specific data, can improve practicality for large-scale speech applications. 

   

\subsection{Mitigating Adversarial Exploitation}
Unlearning can introduce vulnerabilities that adversaries may exploit to manipulate speech systems. Robust defenses are needed against adversarial attacks, such as backdoor unlearning, where an attacker forces selective forgetting to degrade model performance on targeted speech categories. In addition, methods to detect and prevent data poisoning during the unlearning process are crucial to maintaining model integrity. 


\section{Conclusion}
We introduce \textbf{speech unlearning}, a novel and underexplored challenge in AI that focuses on selectively removing a subset of speech data and its influence from trained models, which is crucial for privacy preservation, ethical AI, and regulatory compliance. While machine unlearning has been investigated in domains such as vision, language models, and recommendation systems, its application to speech has unique challenges due to the the high-dimensional, sequential, and speaker-dependent nature of speech data. We investigate speech unlearning in two fundamental speech tasks (keyword spotting and speaker identification) and discussed key challenges such as feature entanglement, temporal dependencies, and the trade-off between unlearning and model utility.  
Our results show that current unlearning methods can't effectively remove learned speech information without compromising model performance, and that a structured training strategy based on curriculum learning could be a promising solution. To address these challenges, future research should investigate new paradigms, such as feature-level unlearning, scalable forgetting strategies, and provable guarantees for efficient and verifiable speech unlearning.







\bibliographystyle{IEEEtran}
\bibliography{mybib}

\end{document}